\documentclass[conf]{new-aiaa}
\usepackage[utf8]{inputenc}

\usepackage{graphicx}
\usepackage{subcaption}
\usepackage{amsmath}
\usepackage[version=4]{mhchem}
\usepackage{siunitx}
\usepackage{longtable,tabularx}
\title{A Unified Generative-Predictive Framework for Deterministic Inverse Design}

\author{Reza T. Batley\footnote{Graduate Research Assistant, Kevin T. Crofton Department of Ocean and Aerospace Engineering} and Sourav Saha\footnote{Assistant Professor, Kevin T. Crofton Department of Aerospace and Ocean Engineering, 1600 Innovation Drive, AIAA Young Professional Member.}}
\affil{Kevin T. Crofton Department of Aerospace and Ocean Engineering, Virginia Polytechnic Institute and State University, 1600 Innovation Drive, Blacksburg, VA, 24060, USA}

\begin{document}

\maketitle

\begin{abstract}
Inverse design of heterogeneous material microstructures is a fundamentally ill-posed and famously computationally expensive problem. This is exacerbated by the high-dimensional design spaces associated with finely resolved images, multimodal input property streams, and a highly nonlinear forward physics. Whilst modern generative models excel at accurately modeling such complex forward behavior, most of them are not intrinsically structured to support fast, stable \emph{deterministic} inversion with a physics-informed bias. This work introduces Janus, a unified generative-predictive framework to address this problem. Janus couples a deep encoder-decoder architecture with a predictive KHRONOS head, a separable neural architecture. Topologically speaking, Janus learns a latent manifold simultaneously isometric for generative inversion and pruned for physical prediction; the joint objective inducing \emph{disentanglement} of the latent space. Janus is first validated on the MNIST dataset, demonstrating high-fidelity reconstruction, accurate classification and diverse generative inversion of all ten target classes. It is then applied to the inverse design of heterogeneous microstructures labeled with thermal conductivity. It achieves a forward prediction accuracy $R^2=0.98$ (2\% relative error) and sub-5\% pixelwise reconstruction error. Inverse solutions satisfy target properties to within $1\%$ relative error. Inverting a sweep through properties reveal smooth traversal of the latent manifold, and UMAP visualization confirms the emergence of a low-dimensional, disentangled manifold. By unifying prediction and generation within a single latent space, Janus enables real-time, physics-informed inverse microstructure generation at a lower computational cost typically associated with classical optimization-based approaches.
\end{abstract}

\section{Nomenclature}

{\renewcommand\arraystretch{1.0}
\noindent\begin{longtable*}{@{}l @{\quad=\quad} l@{}}
$\mathcal{L}$ & Loss \\
$\mathcal{L}_\mathrm{entropy}$ & Cross-Entropy Loss \\
$\mathcal{L}_\mathrm{recon}$ & Reconstruction Loss \\
$\mathcal{L}_\mathrm{cycle}$ & Cycle Consistency Loss \\
$\mathcal{L}_\mathrm{deep}$ & Deep Cycle Consistency Loss \\
$\mathcal{L}_\mathrm{align}$ & Manifold Alignment Loss \\
$R^2$ & Coefficient of Determination \\
RMSE & Root Mean-Square Error \\
$\mathcal{E}$ & Encoder \\
$\mathcal{D}$ & Decoder \\
$\mathcal{K}$ & Predictive Head \\
$\mathcal{X}$ & Input Space \\
$\mathcal{Y}$ & Output Space \\
$\mathcal{Z}$ & Latent Space \\

\end{longtable*}}

\section{Introduction}

\lettrine{R}{ecent} advances in large-scale visual models, encompassing Vision Transformers (ViTs) \cite{dosovitskiy2021image}, vision-language models (VLMs) \cite{radford2021learning}, generative adversarial networks (GANs) \cite{goodfellow2014generative} and diffusion models \cite{rombach2022high}, have led to systems capable of synthesising high-fidelity imagery from latent encodings. Yet, despite their revolutionary generative power, these architectures remain primarily optimized for the \textit{forward generative} direction. Their latent spaces optimize for sampling, impressionistic guidance, and probabilistic decoding, not for \emph{deterministic, cycle-consistent inversion}. For some desired output, whether it is an aerodynamic coefficient, a target classification, or an image, there is no general, principled mechanism to recover a latent representation both consistent with the desired output and stable under the model's own forward-inverse dynamics. Inverse queries are approximated, rather, through heuristic prompt-chaining or latent search procedures, stochastic sampling or re-optimization of the latent space in a way not aligned with the forward task. The present work addresses this limitation. \emph{Janus} learns a latent manifold $\mathcal{Z}$ onto which raw, multimodal inputs are mapped via its encoder $\mathcal{E}$, in a way not dissimilar to an autoencoder \cite{hinton2006reducing}. This latent manifold is then fed as input into Janus' two heads. One is its decoder $\mathcal{D}$, and the other its predictive head $\mathcal{K}$. In fact, $\mathcal{K}$ takes the form of a KHRONOS head \cite{batley2025khronos}. The \textbf{key novelty} of Janus lies in that this manifold is jointly optimized both for cycle-consistency (via $\mathcal{E}$ and $\mathcal{D}$) and predictive accuracy (via $\mathcal{K}$).

Formally, let $\mathcal{X}$ denote an input space and $\mathcal{Y}$ an output space. Concretely, Janus encodes inputs via $\mathcal{E}:\mathcal{X}\rightarrow \mathcal{Z}$, maps this to task outputs with $\mathcal{K}:\mathcal{Z}\rightarrow \mathcal{Y}$ and decodes via $\mathcal{D}:\mathcal{Z}\rightarrow \mathcal{X}$. The forward model is then, in effect, $\mathcal{K}\circ\mathcal{E}:\mathcal{X}\rightarrow\mathcal{Y}$. This model is trained to minimize an appropriate loss functional, as discussed in greater detail in Section \ref{sec:loss}. Successful training shapes $\mathcal{Z}$ into a geometry that is a) readily decodable and b) information rich for the predictive head. If $\mathcal{K}$ is then chosen to be readily invertible such that one may find $\mathcal{K}^{-1}:\mathcal{Y}\rightarrow\mathcal{Z}$, Janus enables end-to-end \emph{generative inversion} via $\mathcal{D}\circ\mathcal{K}^{-1}:\mathcal{Y}\rightarrow\mathcal{X}$.

In what follows, this framework is first instantiated in the form of a concrete architecture known as \emph{Janus-C}. In this model, the encoder $\mathcal{E}$ and decoder $\mathcal{D}$ are realized as convolutional modules. To demonstrate Janus' architectural generality, a brief introduction of \emph{Janus-ViT} is included. Here, $\mathcal{E}$ is a Vision Transformer and $\mathcal{D}$ combines a transformer feeding a bilinear upscale-convolution model, this decoder setup designed as in \cite{esser2021taming}.

This makes the following three key contributions:
\begin{itemize}
    \item The introduction of \textbf{Janus}, a conceptual framework in which a single latent manifold is jointly optimized to support forward-inverse consistency and predictive performance;
    \item Demonstration that the resultant latent space admits rapid inversion at a cost on the same order as a single forward inference; and
    \item Introduction of two instantiations of Janus, Janus-C and Janus-ViT. Janus-C is thoroughly validated for generative inverse design of microstructure from a desired set of target properties.
\end{itemize}

\begin{figure}[h!]
    \centering
    \includegraphics[width=\linewidth]{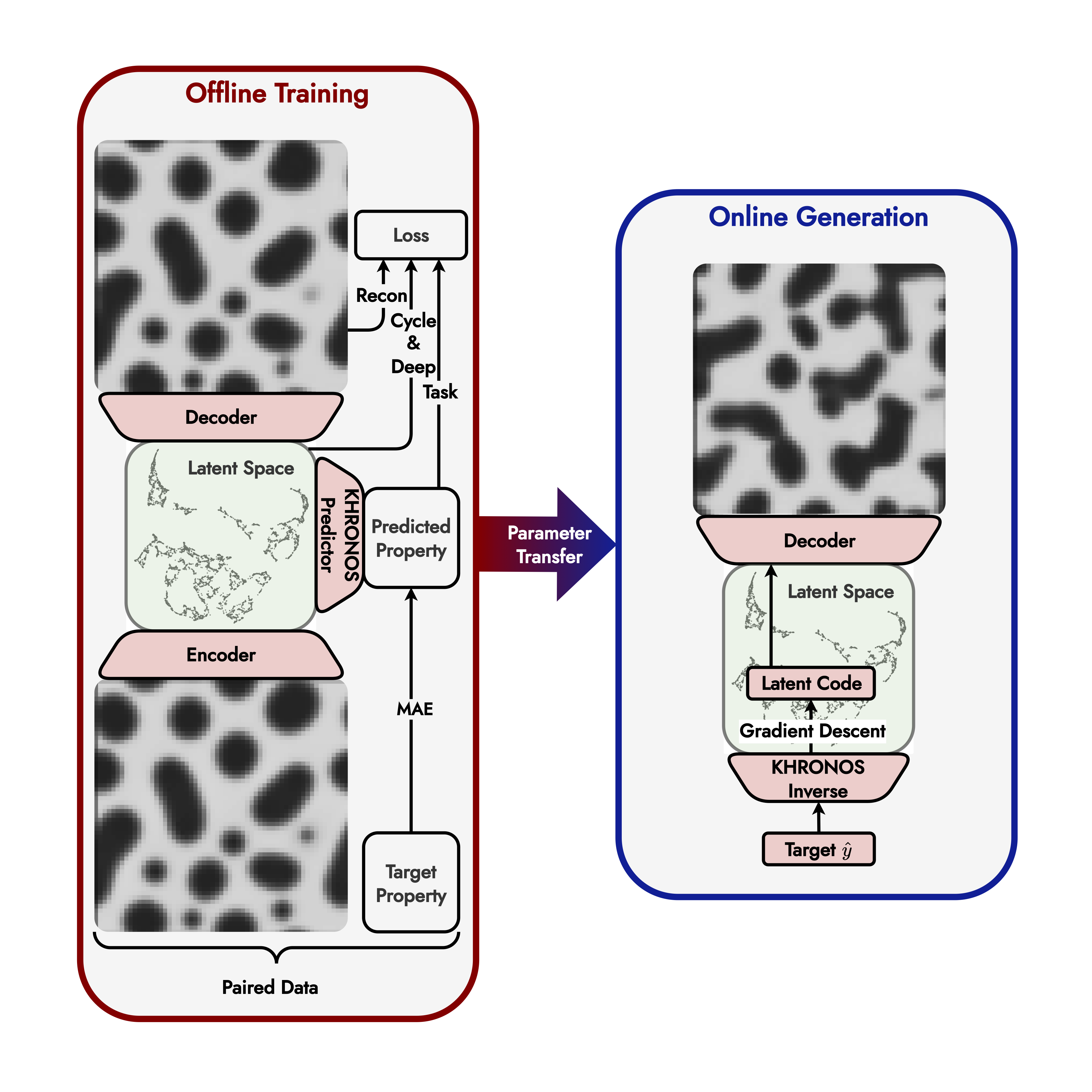}
    \caption{Overview of Janus' framework for inverse design of microstructure. During offline training (\emph{left}), paired microstructure-property data are encoded into a jointly optimized shared latent space. Learned parameters are transferred into the online generation stage (\emph{right}), where a target property is inversely mapped back to a latent code in the space. This code is then decoded by the trained decoder, resulting in a deterministic inverse generation.}
    \label{fig:overview}
\end{figure}

\noindent A schematic overview of the Janus training and generative inversion process is provided in Figure \ref{fig:overview}. The efficacy of the framework is tested against benchmark experiments (MNIST datasets). The key scientific contribution of this paper is in the application of Janus trained on a phase-field generated microstructure dataset. The goal is to generatively invert a target property to a heterogeneous microstructure yielding said desired property. In this work, this property is thermal conductivity. 

\section{Background}
\label{sec:background}
\subsection{Separable Neural Architectures (SNA)}
Separable Neural Architectures (SNAs), as introduced formally in \cite{batley2025explainable}, and building upon work in \cite{park2025unifying,saha2024mechanistic,saha2021hierarchical}, represent $d$-dimensional functions by compositions of low-arity, learnable atoms $\phi^{(S)}:[0,1]^{|S|}\times\Theta_S\rightarrow\mathbb{R}$ operating, without loss of generality, on the unit hypercube. $\Theta_S$ is the space of learnable parameters, and $S\subset[d]$ is a subset of the index set $[d]=\{1,\dots,d\}$. The \emph{interaction order} $k$ bounds dimensional interactivity by setting $|S|\leq k$. The composition itself is dictated by a finite-rank interaction object $\mathcal{C}=\{c_S:S\subset [d], S\leq k\}$. This rank constraint is controlled by the \emph{rank-restriction} hyperparameter $r$, so that $\operatorname{rank}(C)\leq r$. This defines the functional representation as,
\begin{align}
    f(x)=\sum_{|S|\leq k}c_S\phi^{(S)}(x_S;\theta_S),~\theta_S\in\Theta_S,
\end{align}
where $x_S$ is the restriction of $x$ to coordinates in $S$.

This separable formulation yields models that are compact, data-efficient and well-conditioned when the to-be-learned mapping itself possesses low intrinstic dimensionality.
\subsection{KHRONOS}
KHRONOS is an instance of SNA. In particular, its structure takes the form of a tensor product between its unary atoms. When tasked to approximate appropriately structured relationships - \textit{i.e.}, those of low intrinsic dimensionality - as is typical in scientific modeling, KHRONOS is exceptionally efficient. This structure leads to models that are both compact and well-conditioned. Each unary atom is typically chosen to be a B-spline basis expansion. 

KHRONOS excels when operating on data already residing in a compact, structured vector space representation. It does not, however, provide a clear mechanism for constructing such representations from high-dimensional raw inputs such as images, natural language, or multimodal data. It is this limitation that Janus is designed to address. It provides an appropriate encoder $\mathcal{E}$ to project all inputs to a KHRONOS-compatible, even KHRONOS-\emph{optimized} latent space. One that is richly predictive, reconstructively accurate via its decoder $\mathcal{D}$, and deterministically cycle-consistent.

\subsection{Latent Generative Models}
Latent generative models construct lower-dimensional representations of data from which high-fidelity samples can be synthesized. Differences between such models lie chiefly in how said latent space is structured and how decoding is performed, leading to tradeoffs in expressivity and stability.
\paragraph{Variational Autoencoders (VAEs).}
VAEs \cite{kingma2022auto,walker2024unsupervised} learn a probabilistic latent representation by penalizing deviations from the latent distribution to that of a sampled prior. Their latent spaces, however, are not cycle consistent. Typically, encoding both an input and its reconstruction would not lead to the same latent representation.
\paragraph{Generative Adversarial Networks (GANs).}
GANs synthesize data through an adversarial "game" between a generator and a discriminator. Variants, such as StyleGAN \cite{karras2019style} , may produce highly structured latent spaces and produce sharp imagery. Encoder-GAN hybrids \cite{donahue2017adversarial} can also improve inversion, however, iterative optimization or separate encoder networks are required. These also do not guarantee stability between forward and inverse passes.
\paragraph{Vector-Quantized Models.}
Vector quantized autoencoders (VQ-VAEs) \cite{oord2018neural} and GANs (VQ-GANs) \cite{esser2021taming} discretize the latent space into a learned embedding codebook. This yields a stable and compact space of latent representations that pair well with transformers and auto-regressive models to model global spatial dependencies in the latent domain rather than pixel space. This makes them particularly capable of synthesizing high-resolution imagery. The quantization step, whilst well-optimized for generative convenience, is non-differentiable with codebook lookup dynamics that complicate true end-to-end inversion.

\paragraph{Diffusion Models.} Diffusion models \cite{rombach2022high, ho2020denoising} synthesize by reversing a learned process of corruption to noise. This model class represents the current state-of-the-art in generative fidelity and support many conditioning mechanisms. In application to generative inversion, however, diffusion models require running deterministic reverse-diffusion or a massively expensive gradient-based latent search. These models do not naturally yield a stable latent representation useful not only for reconstruction but also prediction. \\
\\
Janus \textbf{strategically differs} from these approaches in ways as follows: the latent space is jointly optimized to be a) decodable, b) predictive, and c) cycle-consistent to enable direct, efficient and reliable inversion without iterative search of the gargantuan parameter space of contemporary generative models.

\subsection{Invertible Models}
Invertible and cycle-consistent architectures seek to ensure learned, forward mappings are reversible. These models, whilst establishing important foundations and inspiration for bidirectional modeling, differ in key respects from the goals of Janus.
\paragraph{CycleGAN.}
CycleGANs \cite{zhu2020cyclegan} enforce cycle-consistency between a pair of image domains - allowing for bidirectional translation without paired supervision. However, the constraint of cycle-consistency is imposed between image spaces, not in a latent space. Further, learned embeddings are not structured to support predictive tasks - leaving them infeasible for most scientific purposes.
\paragraph{Bidirectional Generative Adversarial Networks (BiGANs).}
BiGANs \cite{donahue2017adversarial} equip a standard GAN with an encoder in such a way that both real and generated data share a common latent. This allows approximate latent inversion, but consistency arises from adversarial matching rather than explicit reconstruction or or cycle constraints. As a result, the recovered latent representation is not guaranteed to be stable under forward-inverse composition. Predictive accuracy is not optimized, a drawback rendering BiGANs impractical for scientific purposes.
\paragraph{Invertible Neural Networks and Normalizing Flows.}
Invertible Neural Networks \cite{ardizzone2019inverse} and Normalizing Flows \cite{rezende2016variational, kobyzev2021normalizing} alike provide an interesting alternative. These models construct inherently bijective neural mappings with tractable Jacobians, and are thus inherently invertible by design. This is a powerful feature, and likely better suited to simple problems than Janus, but with a significant caveat. The enforced bijectivity of these models is restrictive in inherently ill-posed for inversion, many-to-one problems. This compresses clearly distinct inputs into nearly identical latent representations - a severe information bottleneck that induces instability, loss of variation and detail. Whilst theoretical universal approximators, this manifests in reality by requiring deep flows in dealing with data in the form of images. These are typically more resource intensive than an equivalently powerful convolutional neural network (CNN), not to mention the problems of vanishing and exploding gradients, slow training and difficult convergence \cite{koehler2021representational}. As such, in practical terms, flows for image-based scientific problems tend to be both more resource-intensive and sensitive to optimize than non-invertible alternatives.
\subsection{Scientific Predictive Surrogate Models}
Scientific surrogate models seek to approximate the forward map, in our case from structure to property. Physics-Informed Neural Networks (PINNs) \cite{raissi2019physics} incorporate, if know, governing equations directly into the loss to improve physical consistency. These can be inverted via external optimization over the input space, but this space is, again, gargantuan. For a scientific problem with thousands or perhaps millions of input parameters, gradient search becomes expensive and sensitive. Approaches of operator learning, such as DeepONets \cite{lu2021deep} and Fourier Neural Operators \cite{li2021fno} learn mappings between function spaces. These methods achieve excellent forward predictive accuracy, but do not yield latent spaces suitable for inversion, much less \emph{optimized} for it. Recovering an input from a desired output requires the same, costly, sensitive iterative search of the full input space. As a result, direct application of these methods involve only forward predictive structure-property linkages \cite{huang2023introduction,mandl2025separable}.

\subsection{Inverse Design and Property-Driven Generation}
Design of aerospace structures often transcends a single length scale. With the advent of additive manufacturing, microstructure design based on a set of target structural properties has become realistic \cite{wang2022multi,wang2025multiscale,billah2025uncertainty}. Classically, inverse design of microstructure has been approached through mechanistic or optimization-based methods. These include topological optimization and level set evolution to iteratively tweak geometry to match a target property. These methods can be effective in continuum-scale design but struggle with microstructures. The inverse of homogenization is an inherently many-to-one problem, so distinct microstructures can produce materials of similar properties. This is not inherently problematic, per se, but rather results in an optimization landscape that is highly non-convex scattered with local minima \cite{gibiansky1997design, torquato2002random}.

Classical surrogate-assisted methods such as Bayesian optimization and surrogate-augmented genetic algorithms attempt to search this space using a predictive forward model \cite{lookman2019active, kalidindi2015materials}. Once the design space is represented by high-dimensional inputs, such as images, the search once again collapses into prohibitively expensive optimization in a gargantuan input space. This brings us to machine learning based inverse design. When using machine learning, inverse problems in materials and design are typically approached either through a) gradient-based optimization in the input space; b) conditional generative models; or c) a latent space search over pretrained models. The philosophy of Janus stems from c), and builds upon it by crafting the latent space itself to be decodable, richly predictive and cycle-consistent. This is moreover achieved whilst ensuring the latent construct remains sufficiently compact to permit its search at effectively negligible computational expense.

\section{Methods}
\subsection{The Foundation: KHRONOS}
In Janus, the predictive head $\mathcal{K}$ is instantiated as a KHRONOS surrogate of quadratic B-spline basis. Concretely, it represents the map $\mathcal{K}:\mathcal{Z}\rightarrow\mathcal{Y}$ using a separable, rank-$r$ expansion of unary basis functions,
\begin{align}
    \mathcal{K}(z)=\sum_{j=1}^r\prod_{i=1}^M\sum_s \alpha_{sij}\psi_{sij}(z_i),\quad z=(z_1,\dots,z_M)\in\mathcal{Z}
\end{align}
where each $\psi$ is a quadratic B-spline defined over a uniform knot vector in $[0,1]$. Further, $M=\dim(\mathcal{Z})$ is the dimension of the latent space chosen such that $M\ll N$ for $N=\dim(\mathcal{X})$ the dimension of the input space.

This representation yields a predictive map that is compact and continuously differentiable. Gradients $\frac{\partial{\mathcal{K}}}{\partial{z}}$, therefore, are stable and inexpensive to compute. This property is essential for the inversion process. Since $\mathcal{K}$ is jointly trained alongside both the encoder $\mathcal{E}$ and decoder $\mathcal{D}$, the latent space is shaped to be maximally rich for prediction while remaining decodable and cycle-consistent.
\subsection{The Extension: Janus}
Janus extends KHRONOS by introducing the aforementioned encoder-decoder pair. The encoder $\mathcal{E}:\mathcal{X}\rightarrow\mathcal{Z}$ maps raw, high-dimensional, potentially multimodal data into the sought latent representation. The decoder $\mathcal{D}:\mathcal{Z}\rightarrow\mathcal{X}$ reconstructs said inputs from the latent space. The predictive KHRONOS head $\mathcal{K}:\mathcal{Z}\rightarrow\mathcal{Y}$ runs the required predictive analysis, and, through its inversion, acts as the entry point for generation. 

Joint training of $\mathcal{E},\mathcal{D}$ and $\mathcal{K}$ shapes the latent manifold to be simultaneously a) decodable; b) predictive; and c) cycle-consistent. Herein lies the key distinction: Janus's latent space is optimized explicitly for inversion, not merely for reconstruction or synthesis.
\subsubsection{Janus-C}
\begin{figure}
    \centering
    \includegraphics[width=\linewidth]{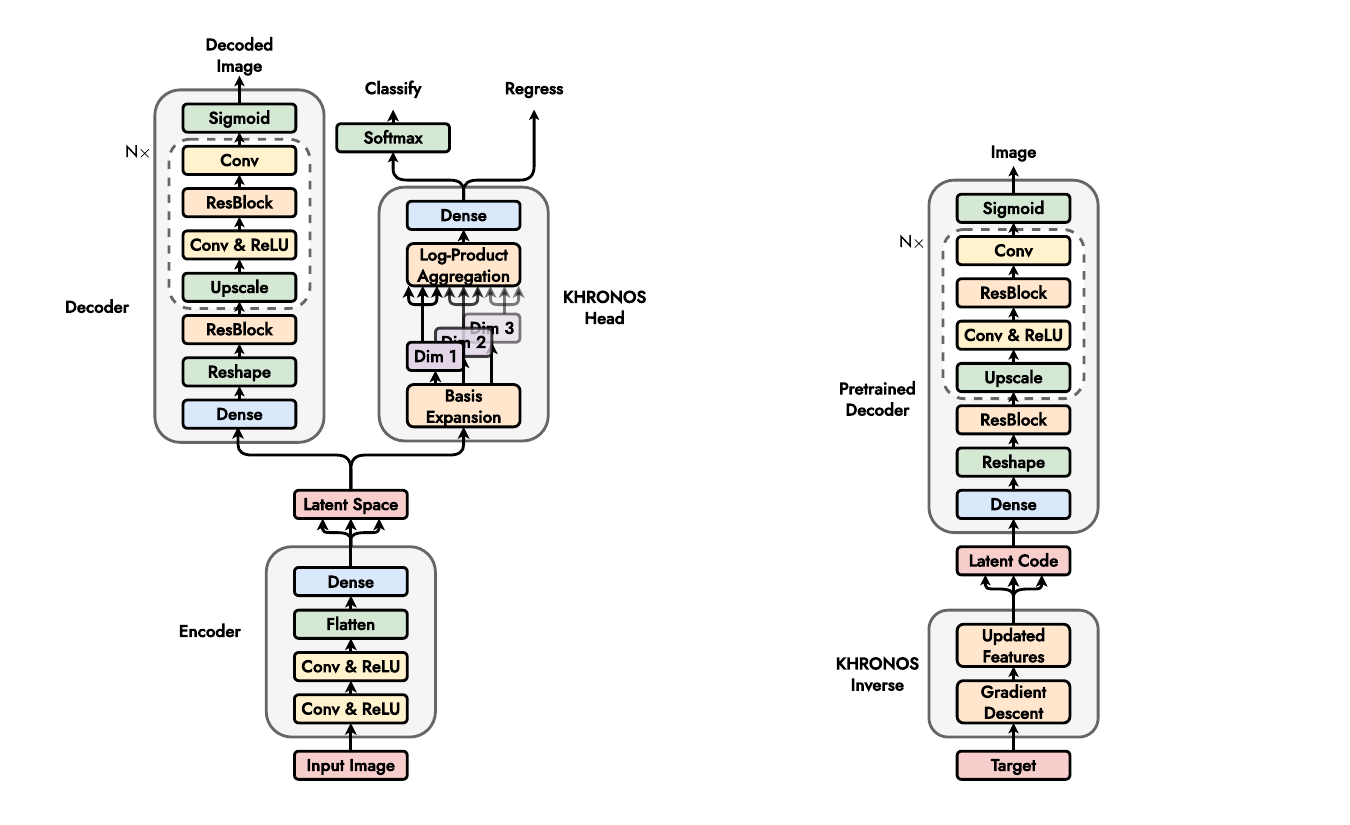}
    \caption{Forward (\textit{left}) and inverse (\textit{right}) architectures of Janus-C.}
    \label{fig:janusc}
\end{figure}

Employing a U-net \cite{ronneberger2015unet} style convolutional encoder-decoder pair results in Janus-C. One writes,
\begin{align}
    \mathcal{E}:\mathcal{X}\xrightarrow[]{\textrm{Conv Encoder}}\mathcal{Z}\quad \mathcal{D}:\mathcal{Z}\xrightarrow[]{\textrm{Conv Decoder}}\mathcal{X}.
\end{align}
The convolutional encoder progressively reduces spatial information and increases channel depth to yield a compact latent embedding. The decoder mirrors this process by repeated bilinear upscaling and convolving. This is visualized in more detail in Figure \ref{fig:janusc} (\emph{left}). This architecture excels, as one might expect, for locally structured image representations. 

\subsubsection{Janus-ViT}
Janus-ViT uses a Vision Transformer to encode images in a way that is more suited to capturing global spatial dependencies,
\begin{align}
    \mathcal{E}:\mathcal{X}\xrightarrow[]{\textrm{ViT Encoder}}\mathcal{Z}\quad \mathcal{D}:\mathcal{Z}\xrightarrow[]{\textrm{Transformer+Conv Decoder}}\mathcal{X}.
\end{align}
The decoder is hybrid. First, latents are embedded into a sequence of spatial tokens. These are given learned positional embeddings which are fed into a transformer. This token sequence is reshaped into a low-resolution feature map before being bilinearly upscaled and convolved to reconstruct a full-resolution image. This design preserves the ViT's advantages in capturing long range interactivity while taking advantage of convolutional refinement for full-fidelity spatial reconstruction. The use of a transformer for modeling long-range relationships before upscaling and refining with convolutions is largely akin to the generator used in \cite{esser2021taming}. 

\subsubsection{The Loss Functional}
\label{sec:loss}
The training objective is thus formulated as a composite loss functional,
\begin{align}
    \mathcal{L}=\mathcal{L}_\mathrm{task}+\lambda_\mathrm{recon}\mathcal{L}_\mathrm{recon}+\lambda_\mathrm{cycle}\mathcal{L}_\mathrm{cycle}+\lambda_\mathrm{deep}\mathcal{L}_\mathrm{deep},
\end{align}
where each $\lambda_{(\cdot)}$ denotes a nonnegative weighting coefficient that balances contributions for each loss term. The "task" loss is the loss guiding the predictive head, typically either as a classification loss, such as cross-entropy, or a regression loss, such as mean-square error.
\paragraph{Reconstruction Loss}
The reconstruction loss is a measure of the decodability of the latent space. Formally, one writes
\begin{align}
    \mathcal{L}_\mathrm{recon}=\mathbb{E}_{x\sim p_\mathrm{data}}\left[\|\mathcal{D}\circ\mathcal{E}(x) - x\|_1\right].
\end{align}
This represents the deviation from input $x$ of a first-order reconstruction.
\paragraph{Cycle Consistency Loss}
The cycle consistency loss is a measure of the reversibility of forward-inverse mappings. This takes the form
\begin{align}
    \mathcal{L}_\mathrm{cycle}=\mathbb{E}_{z\sim p_z}\left[\|\mathcal{E}\circ\mathcal{D}(z) - z\|_1\right]
\end{align}
for an encoded input $z=\mathcal{E}(x)$, $p_z$ denoting the empirical distribution of latent vectors produced by the encoder over a training batch. This loss penalizes structural incoherence and prevents degenerate mappings.
\paragraph{Deep Cycle Consistency}
The deep cycle consistency extends the cycle constraint into a second-order representation space, a sort of curvature constraint. This is written
\begin{align}
    \mathcal{L}_\mathrm{deep}=\mathbb{E}_{x\sim p_\mathrm{data}}\left[\|\mathcal{D}\circ\mathcal{E}\circ\mathcal{D}\circ\mathcal{E}(x) - x\|_1\right].
\end{align}
This loss encourages cycle stability and has neurobiological parallels, discussed later.
\subsection{Generative Inversion}
Generative inversion is the process of generating one or a set of inputs corresponding to some queried target output $\hat y$. This proceeds first by inverting KHRONOS to find that latent submanifold $S_z\subseteq \mathcal{Z}$ each of whose elements recover the target. Formally, denoting the KHRONOS inversion step as $\mathcal{K}^{-1}$, one writes,
\begin{align}
    \mathcal{Z}_{\hat y}=\left\{z\in \mathcal{Z}~|~\mathcal{K}(z)=\hat y\right\}.
\end{align}
Each $\hat z\in\mathcal{Z}_{\hat y}$ can then be decoded to reconstruct a corresponding input $\hat x=\mathcal{D}(\hat z)\in\mathcal{X}$. This recovers the input manifold of interest,
\begin{align}
    \mathcal{X}_{\hat y}=\left\{\mathcal{D}(z)~|~z\in\mathcal{Z}_{\hat y}\right\}.
\end{align}
\paragraph{Manifold Alignment Loss}
The manifold alignment loss acts as a statistical regularizer during inversion, penalizing latent candidates that drift from the learned training distribution. The empirical latent distribution is approximated by its first and second moments $\mu_\mathcal{Z},\sigma_\mathcal{Z}$. The loss is written,
\begin{align}
    \mathcal{L}_\mathrm{align}=\left\|\frac{z-\mu_\mathcal{Z}}{\sigma_\mathcal{Z}}\right\|_2^2.
\end{align}
In minimizing this term, the inverted latent vector remains statistically consistent with the geometry of the encoder's output space. This prevents spurious hallucination.

Having defined the requisite constraints, the generative inversion is executed as a gradient-based optimization process. To recover the target $\hat y$, $\hat z$ is solved for by minimizing the composite inversion objective $\mathcal{J}(z)$. This frames the task as a Maximum a Posteriori (MAP) \cite{gauvain1994maximum} estimation, where the optimizer seeks the most probable latent vector satisfying the property constraint. This objective is given by,
\begin{align}
    \mathcal{J}(z) = \underbrace{\|\mathcal{K}(z) - \hat{y}\|_2^2}_{\text{Task}} + \lambda_\mathrm{align}\mathcal{L}_\mathrm{align}(z) + \lambda_\mathrm{cycle}\mathcal{L}_\mathrm{cycle}(z) + \lambda_\mathrm{deep}\mathcal{L}_\mathrm{deep}(z).
\end{align}
Here the task term simply ensures the latent corresponds to a projection that satisfies the physical property target. The cycle and deep cycle consistencies ensure the latent vector remains in that area of the space where the encoder and decoder remain mutual inverses.

This optimization is randomly initialized with $z_0\sim\mathcal{N}(\mu_\mathcal{Z},\sigma_\mathcal{Z})$. The ill-posed, many-to-one nature of this inverse problem results in different initializations recovering different microstructures. One may then initialize at an array of initial points to return a swarm of candidate inverse microstructures yielding the same target property. 

The vector $z$ is then iteratively updated via Adam iterations to minimize $\mathcal{J}(z)$, sliding the point along the morphologically valid latent manifold until it converges to the target property. The inversion schematic in the Janus-C instance is visualized in Figure \ref{fig:janusc} (\emph{right}).
\section{Results}
\subsection{Validation on Standard Benchmarks: MNIST}
For the sake of clear illustration and to validate the architecture, Janus-C is first applied to the MNIST handwritten digit dataset \cite{lecun1998mnist}. The objective is to verify that joint optimization of predictive accuracy and cycle-consistency yield a latent manifold capable of both targeted and \emph{diverse} generative inversion.
\subsubsection{Forward Competency}
Janus-C is adapted to a arbitrary image domain, here $28\times28$, by scaling spatial dimension of the residual backbone whilst keeping constant decoder head depth and connectivity. The KHRONOS head is configured for classification $\mathcal{Y}\subset\mathbb{R}^{10}$ rather than regression. 

Training stabilizes rapidly, taking only 2 epochs to simultaneously achieve a test classification accuracy above 90\% and pointwise reconstruction fidelity of 10\%. After 50 epochs, the model reaches a test accuracy of 97.5\% and pointwise fidelity around 8\%. This confirms that the chosen latent ($\mathcal{Z}\in\mathbb{R}^{16}$) retained sufficient geometric information both to reconstruct inputs and classify at high accuracy.

\subsubsection{Generative Inversion via Swarm Optimization}
The framework's capacity for inverse design is tested with ``swarm'' optimization. For each target digit $\hat y=0,\dots,9$, a batch of 10 parallel inversions is executed. Each instance is started with a different random seed, and optimized via Adam (learning rate $\alpha=0.01$) for 1,000 steps to minimize the composite objective $\mathcal{J}(z)$. The manifold alignment term uses a conditional prior. Here, $\mu_\mathcal{Z}$ and $\sigma_\mathcal{Z}$ are estimated from a small reference batch of real images belonging to the target class. This choice encourages the optimizer to steer towards the specific latent cluster of each digit.

\begin{figure}[h!]
    \centering
    \includegraphics[width=0.9\linewidth]{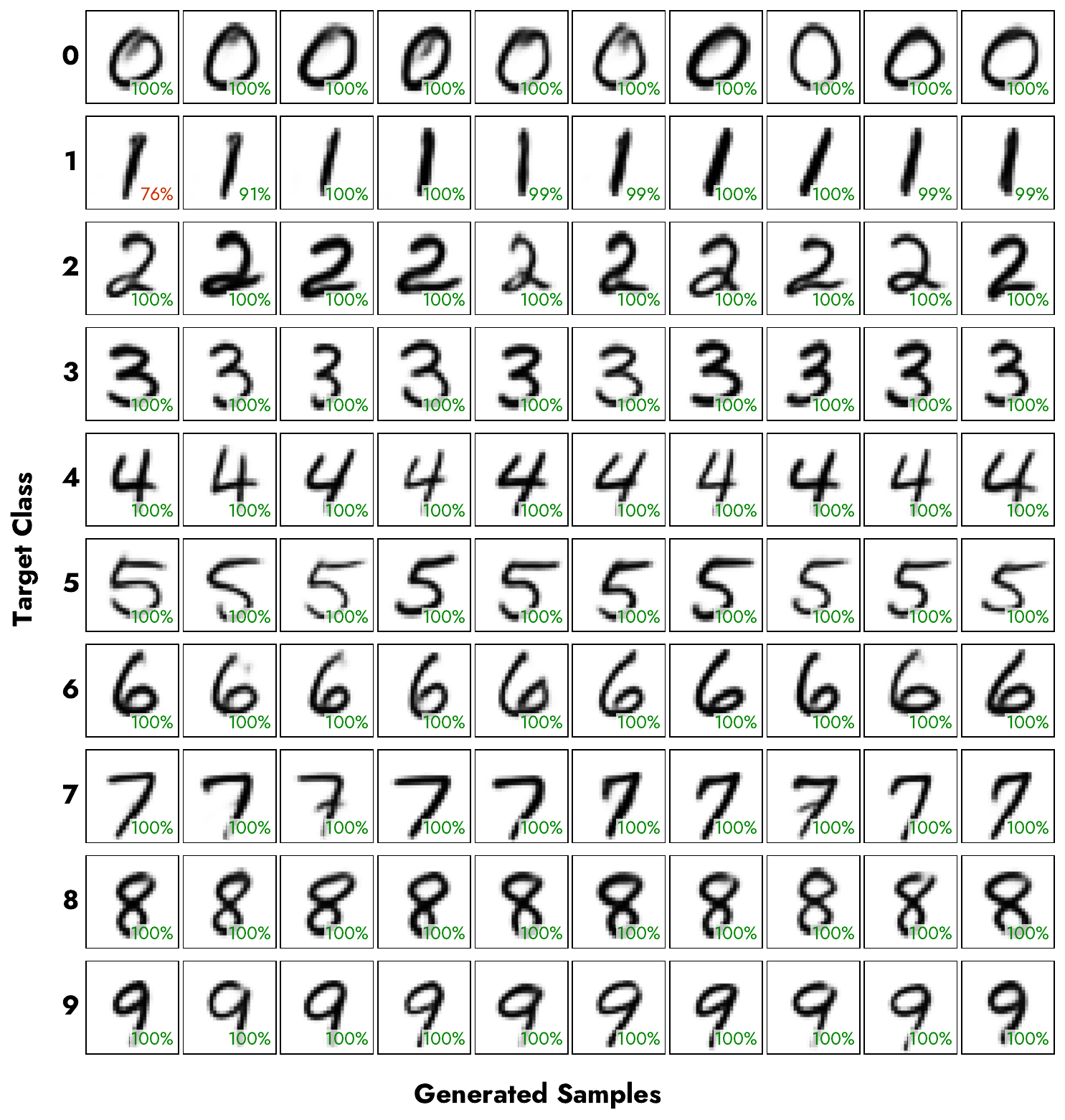}
    \caption{Validation of generative inversion on the MNIST benchmark. Rows correspond to the target digit classes $\{0,\dots,9\}$ and columns display ten independent samples generated via parallel swarm inversion from random latent initalizations. The confidence of the KHRONOS head in each generation is also displayed inset within each.}
    \label{fig:mnist}
\end{figure}

Figure \ref{fig:mnist} illustrates this robust success. The model consistently recovers target digit morphology with high confidence - almost entirely 100\%. The swarm approach highlights the diversity of the solution space with distinct initializations settling into distinct basins of attraction. Notable is the "imperfect" generation of the first sample for the digit "1". Visually, this digit very mildly exhibits a character not unlike a "7". This generation and correspondingly lowered confidence suggest that Janus has learned the continuous boundaries of handwriting styles rather than memorizing discrete representatives. Plainly, the ability to generate and recognize such edge cases points to the smoothness of the learned latent manifold.

\subsection{Application to Generative Inversion of Microstructure}
The framework's capability is here evaluated in a scientific context. In particular, Janus-C is applied to the inverse design of material microstructures. The model is trained on the OSTI dataset \cite{attari2023towards}, containing binarized two-phase microstructures labeled with their effective thermal conductivity $k$. All reported conductivity values are dimensionless, normalized quantities. In particular, this data stems from phase-field based simulations in which a spatially resolved phase variable delineates distinct regions, e.g. solid and pore. This model evolves according to the thermodynamic and kinematic equations to produce realistic as well as statistically and physically plausible microstructure patterns. 

The KHRONOS head is thus configured as a continuous regressor, marking the only structural change from the MNIST run. To ensure numerical stability during training, target values are scaled up by a factor of 10,000. This modification is made without loss of generality, for the provided conductivity data is dimensionless. The objective is again to shape a latent manifold $\mathcal{Z}$ so that it is simultaneously isometric to the geometric input space and smoothly correlated to the physical property space, at least in such a way that it readily feeds the KHRONOS head. 

\subsubsection{Forward Training Dynamics}

\begin{figure}
    \centering
    \includegraphics[width=\linewidth]{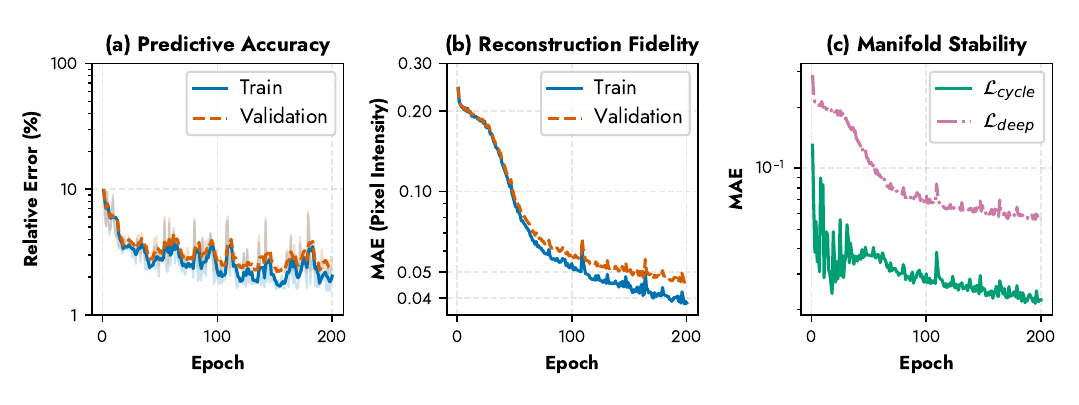}
    \caption{Training dynamics of Janus-C. (a) Regression learning curves with relative accuracy converging to 2.5\%. (b) Reconstructive fidelity, measured by mean absolute error in pixel intensity, decreasing to sub 0.05. (c) Manifold stability in mean absolute terms for cycle-consistency and deep cycle losses.}
    \label{fig:training}
\end{figure}

First assessed is the stability of the multi-objective learning process. As the loss function pulls between both pixel-perfect reconstruction and regressive accuracy, a notable result would be competency in both. Figure \ref{fig:training} illustrates the evolution of the component metrics. As shown in Panel (a), the predictive head converges rapidly to achieve a final validation mean absolute error (MAE) of 0.67, corresponding to a relative error of around 2.5\% and coefficient of determination ($R^2$) exceeding 0.97. This indicates that the encoder successfully extracts those morphological descriptors governing thermal transport within the 64-dimensional latent bottleneck.

At the same time, the reconstruction fidelity, Panel (b), stabilizes at a pixelwise MAE of 0.046, a sub-5\% intensity error. Given the largely binary nature of the material images, this low error magnitude suggests the decoder well-resolves sharp phase boundaries without the "averaging" blurry artifacts associated with under-regularized models. Finally, Panel (c) visually validates the structural integrity of the inversion. The cycle-consistency loss smoothly converges to below 0.03 in absolute terms, and the deep cycle loss tracks the reconstruction baseline closely. This indicates that the latent manifold is stable under re-encoding. 

\subsubsection{Generative Inversion}
\label{sec:geninv}
\begin{figure}[h]
    \centering
    \includegraphics[width=\linewidth]{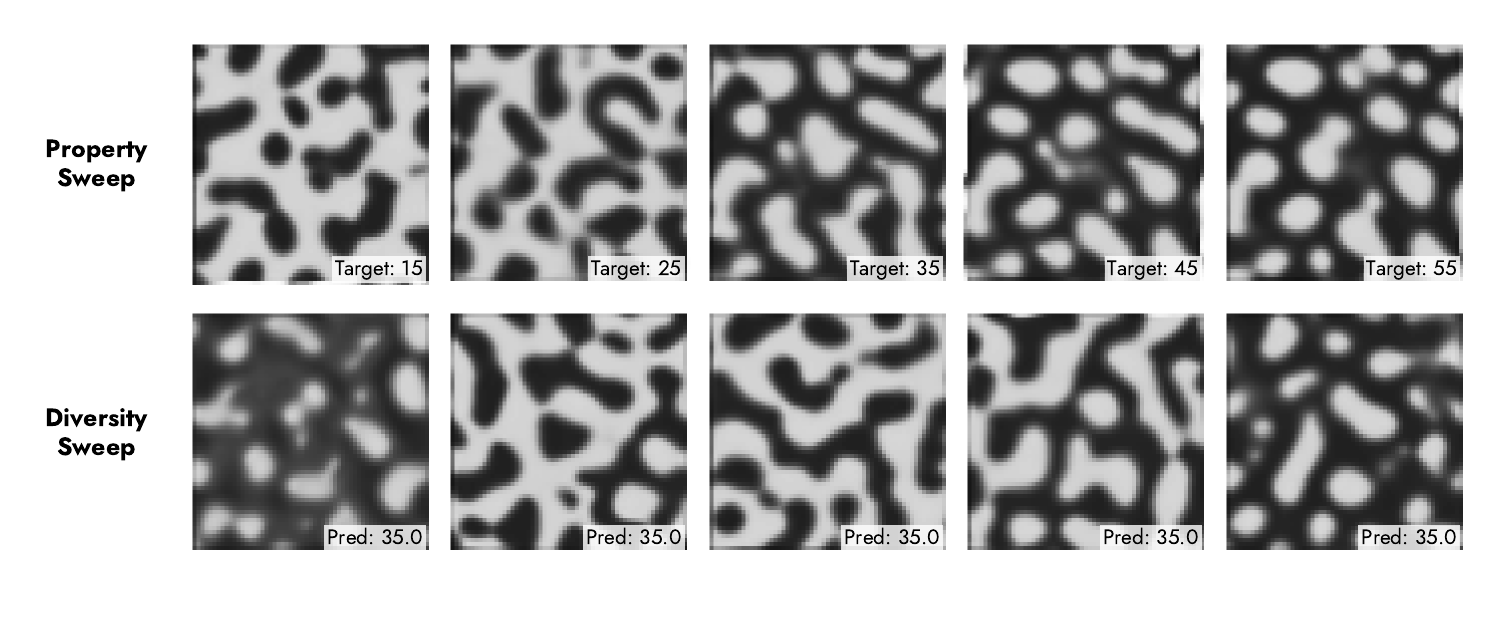}
    \caption{Generative inversion of microstructure. Top row: property sweep demonstrating deterministic targeting of microstructure for thermal conductivities of $k=15,25,35,45$ and $55$ (dimensionless). Bottom row: diversity analysis for a fixed target $35$ starting from five distinct starting seeds.}
    \label{fig:inverse}
\end{figure}

Having established stability of the forward model, the objective is targeted inversion to demonstrate the framework's capability for inverse design. Figure \ref{fig:inverse} presents the results of two distinct experiments conducted via Maximum a Posteriori (MAP) estimation in the latent space. The top row (\emph{Property Sweep}) visualizes the traversal of the latent manifold along the thermal conductivity gradient. In particular, $k=15,25,35,45$ and $55$ are sequentially targeted. A smooth morphological transition is seen, demonstrating a form of topological ordering with respect to the thermal conductivity of the latent space. 

The bottom row (\emph{Diversity Sweep}) addresses the ill-posedness of inverse microstructure design. For a fixed target of $35$, inversion from five random initial seeds yield five distinct morphologies. Despite clear visual differences, all five generations match the target property to well under 1\% relative error, according to the KHRONOS regressor head. Janus-C effectively captures the nullspace of the inverse problem.

\begin{figure}
    \centering
    \includegraphics[width=\linewidth]{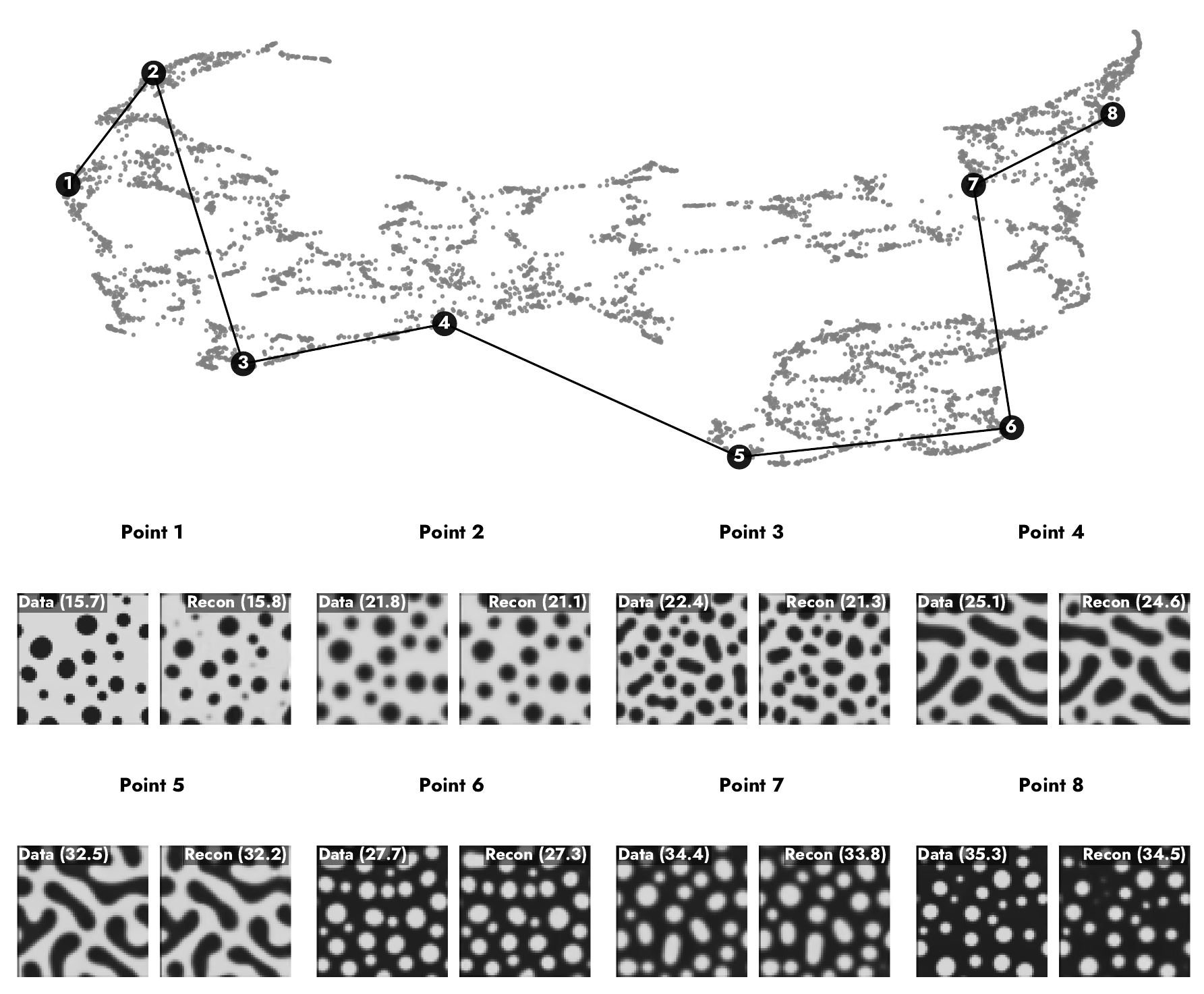}
    \caption{A latent space traversal in the Janus-C manifold. This continuous trajectory is generated by ordering latent codes along the dominant direction of identified variation via PCA. The numbered points denote locations sampled along this trajectory. At each position, the corresponding dataset microstructure (\emph{Data}) with its paired thermal conductivity and its decoded reconstruction (\emph{Recon}) with its predicted thermal conductivity are shown.}
    \label{fig:traverse}
\end{figure}

A further view of the learned latent geometry is provided in Figure \ref{fig:traverse}. Here a continuous traversal of the learned manifold is obtained by ordering encoded dataset samples along the dominant principal component. These are overlaid on the space's Uniform Manifold Approximation and Projection (UMAP) projection \cite{mcinnes2020umap}. Each of these latent codes is decoded by Janus' trained decoder head and presented beside its original sample. The progression along this path forms a strikingly smooth and monotonic morphological continuum starting from predominantly white fields with sparse black inclusions. These inclusions become denser then morph into filament structures, eventually ending in predominantly black regions with sparse white inclusions. This visual monotonic progression is tracked to some extent by each code's thermal conductivity values (both ground truth and predicted), bar an outlier at Point 5.

This monotonic sweep reflects that the learned latent manifold has organized such that the dominant axes correspond both to physical as well as topologically significant modes of variation. This is in stark contrast to VAEs whose latent manifolds entangle many unrelated generative factors, typically resulting in a distinct blob-like cloud. A trivial corollary is that movements in principal directions correspond to smooth, coherent changes in topology. This is precisely the behavior a useful invertible surrogate model requires. 

\section{Discussion}
\subsection{Interpretation of Latent Geometry}
Janus' efficacy arises in large part due to specific topological constraints imposed on the latent manifold $\mathcal{Z}$. In the standard generative models, as discussed in length in Section \ref{sec:background}, the latent space is shaped by probabilistic priors or an adversarial discriminator. This ensures that sampled points decode to realistic images, but not that the space is well-structured for deterministic inverse design. Janus, in contrast, enforces a dual-objective topology. The predictive head $\mathcal{K}$ exerts a "stringing" effect organizing the latent space in such a way that trajectories in $\mathcal{Z}$ correlate smoothly with the target property $y$. The cycle-consistency loss $\mathcal{L}_\mathrm{cycle}$ acts as geometric regularizer. This ensures the mapping $\mathcal{E}:\mathcal{X}\rightarrow\mathcal{Z}$ acts as an approximate isometry on the data manifold.

\begin{figure}[h!]
    \centering
    \includegraphics[width=0.75\linewidth]{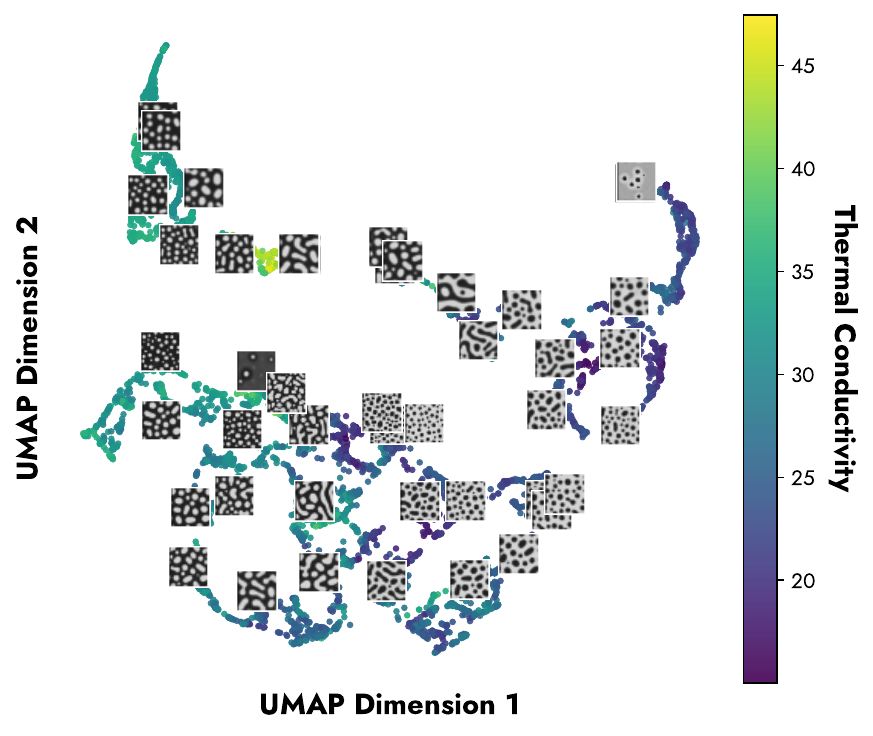}
    \caption{Visualization of the $64$-dimensional latent manifold $\mathcal{Z}$ via UMAP. Points are colored according to their thermal conductivity $k$. Overlaid are representative microstructures from the training distribution mapped to their projected positions in the latent space.}
    \label{fig:umap}
\end{figure}

This topological structuring is visually identifiable in Figure \ref{fig:umap}. The UMAP diagram reveals the latent distribution does not form an isotropic Gaussian cloud, standard in Kullback-Leibler regularized models, rather organized into distinct, continuous strings of smoothly varying conductivity. This topology suggests that Janus has found and preserved a low intrinsic dimensionality manifold of physically valid microstructures.

The \emph{property sweep} generative inverse process carried out in Section \ref{sec:geninv} can thus be understood as traversing one of these strings. Thus, the inversion of KHRONOS only requires moving along these strings, not even a search of the entire latent space. The absence of discontinuities complements the high stability and physical validity of the generated designs, as the inverse optimizer is guided away from invalid regions, moving only along strings.

This structure explains the rapid convergence seen in Figure \ref{fig:training} by effectively collapsing the nullspace, if you will, of the decoder. Regions of the latent space that do not map to physically realizable structures are pruned by the cycle-consistency constraints. In standard autoencoders, the decoder tends to learn a many-to-one mapping where vast areas of $\mathcal{Z}$ map to generic, blurry, or otherwise physically meaningless outputs. By enforcing $\mathcal{E}(\mathcal{D}(z))\rightarrow z$, Janus ensures that the latent manifold is essentially bijective with the physical data manifold. Enforcement of the same condition in KHRONOS head inversion acts in a similar way via constraining the inverse optimizer to keep to near-isometric regions of the latent space. 

\subsection{Implications of Computational Efficiency}
While the primary motivation for this framework is in the enforcement of both physical and topological consistency, the architecture offers cost advantages over traditional inverse design methods. Classical approaches, such as genetic algorithms and topological optimization, suffer from the curse of dimensionality. They require the evaluation of heavy forward models, such as finite element analysis, over thousands of iterations in a high-dimensional search space \cite{gibiansky1997design}.

In contrast, Janus shifts the computational burden of optimization from $\mathbb{R}^{64\times 64}$ pixel space to the comparatively tiny latent manifold in $\mathbb{R}^{64}$. Not only smaller, this manifold, as discussed, yields a well-conditioned optimization landscape. The differentiable KHRONOS head $\mathcal{K}$ allows for negligible-in-time gradient computation compared to numerical physics solvers. In the inverse design of microstructure experiment, physically valid designs satisfying strict property targets ($<1\%$ relative error) were, without fail, recovered in 2,000 gradient steps. The entire inverse pipeline, completed by feeding the optimized latent code to the decoder, is a single second per-point endeavor on the GPU used, an NVIDIA a100 PCIe 40GB. This capability enables real-time material exploration, a departure from the prohibitive computational costs associated with classical inverse design of microstructure.
\subsection{Limitations and Future Work}
Whilst Janus demonstrates significant potential for accelerating inverse design, some limitations are notable. First, the framework's dependence on the underlying training distribution. Janus is inherently interpolative rather than extrapolative, so the validity of generated microstructures and accuracy of property predictions are not guaranteed outside the data manifold. 

Second, the instance of interest Janus-C relies on a convolutional backbone operating on 2D discretizations. While effective for the reduced-order models presented, scaling to high-resolution, three-dimensional microstructures would impose a greater computational footprint. This may necessitate the adoption of more scalable backbones such as the Vision Transformers (Janus-ViT) outlined in section 3.

Future work might focus on two primary avenues. The KHRONOS predictive head $\mathcal{K}$ may be extended to multi-objective settings. Real-world material design tends to trade-off competing properties, such as increasing thermal conductivity whilst minimizing modulus. A vector-valued KHRONOS head would allow such multi-property objective latent optimization. Janus might also be integrated into an active learning loop. By using Janus to propose novel candidate microstructures that are validated by high-fidelity physics simulations, both the training dataset and supported manifold can be iteratively refined. This can be seen as an analogue of Reinforcement Learning from Human Feedback (RLHF) in LLMs \cite{ouyang2022training}.

\section{Conclusion}

This work introduced Janus, a unified framework for generative inverse design that seeks both high-fidelity representation and accurate predictive physics. By fundamentally rethinking the role of the latent manifold - optimizing for a property-aware geometry rather than a passive receptacle of compressed features - Janus resolves the stability and efficiency bottlenecks traditionally plaguing inverse design.

The results demonstrate that this joint optimization objective indeed collapses the typically isotropic cloud to a topologically ordered set of stable and smoothly varying strings. This intrinsic structure ensures that the typically ill-posed and non-convex inverse problem becomes a well-conditioned gradient traversal in the latent space. Demonstrated both on generating handwritten digits with high confidence and designing microstructures with precise thermal properties, Janus navigates the complex nature of inverse design with low - on the order of a second - computational cost.

Ultimately, Janus represents an ongoing paradigm shift in scientific machine learning away from unconnected black-box surrogates, towards interpretable and geometrically structured manifolds. Plainly, this shift moves generative inverse design away from stochastic search and to deterministic navigation.
\clearpage
\section*{Acknowledgments}
S.Saha gratefully acknowledges the start-up fund provided by the Kevin T. Crofton Department of Aerospace and Ocean Engineering, Virginia Polytechnic Institute and State University. R.T. Batley acknowledges the Crofton Fellowship from the Kevin T. Crofton Department of Aerospace and Ocean Engineering, Virginia Polytechnic Institute and State University.
\bibliography{sample}

\end{document}